\documentclass[sigconf]{acmart}
\AtBeginDocument{%
  \providecommand\BibTeX{{%
    \normalfont B\kern-0.5em{\scshape i\kern-0.25em b}\kern-0.8em\TeX}}}

\copyrightyear{2021}
\acmYear{2021}
\setcopyright{acmcopyright}\acmConference[CIKM '21]{Proceedings of the 30th ACM International Conference on Information and Knowledge Management}{November 1--5, 2021}{Virtual Event, QLD, Australia}
\acmBooktitle{Proceedings of the 30th ACM International Conference on Information and Knowledge Management (CIKM '21), November 1--5, 2021, Virtual Event, QLD, Australia}
\acmPrice{15.00}
\acmDOI{10.1145/3459637.3482204}
\acmISBN{978-1-4503-8446-9/21/11}

\usepackage{amsmath,amsfonts,bm}

\def\eqref#1{equation~\ref{#1}}

\def\1{\bm{1}}

\DeclareMathAlphabet{\mathsfit}{\encodingdefault}{\sfdefault}{m}{sl}
\SetMathAlphabet{\mathsfit}{bold}{\encodingdefault}{\sfdefault}{bx}{n}

\usepackage{hyperref}
\usepackage{url}

\usepackage{microtype}
\usepackage{subcaption}
\usepackage[tableposition=above]{caption}
\usepackage{microtype}
\usepackage{graphicx}
\usepackage{algorithm}
\usepackage{algorithmic}

\usepackage{amsmath,amsfonts,amssymb,amsthm,nccmath}
\usepackage{thmtools,thm-restate}
\usepackage{enumerate}
\usepackage{booktabs} 
\usepackage{multirow}
\usepackage[export]{adjustbox}
\usepackage{bbm}
\usepackage{layouts}
\usepackage{mathtools}
\usepackage{wrapfig}
\usepackage{enumitem}
\usepackage{balance}

\usepackage{hyperref,url}
\def\shownotes{1} 
\ifnum\shownotes=1
\newcommand{\authnote}[2]{{$\ll$\textsf{\footnotesize #1 notes: #2}$\gg$}}
\else
\newcommand{\authnote}[2]{}
\fi

\newcommand{\myparagraph}[1]{\noindent\textbf{#1}}

\begin{document}
\fancyhead{}

\title{Towards Robustness to Label Noise in Text Classification via Noise Modeling}


\author{Siddhant Garg}
\authornote{Equal contribution. Work completed at the University of Wisconsin–Madison.}
\affiliation{%
  \institution{Amazon Alexa AI}
  \country{}
}
\email{sidgarg@amazon.com}

\author{Goutham Ramakrishnan}
\authornotemark[1]
\affiliation{%
  \institution{Health at Scale Corporation}
  \country{}
}
  \email{goutham7r@gmail.com}
  
\author{Varun Thumbe}
\authornotemark[1]
\affiliation{%
  \institution{KLA Corporation}
  \country{}
}
  \email{thumbevarun@gmail.com}  



\begin{abstract}
Large datasets in NLP tend to suffer from noisy labels due to erroneous automatic and human annotation procedures.
We study the problem of text classification with label noise, and aim to capture this noise through an auxiliary noise model over the classifier. 
We first assign a probability score to each training sample of having a clean or noisy label, using a two-component beta mixture model fitted on the training losses at an early epoch.
Using this, we jointly train the classifier and the noise model through a novel de-noising loss having two components: (i) cross-entropy of the noise model prediction with the input label, and (ii) cross-entropy of the classifier prediction with the input label, weighted by the probability of the sample having a clean label.
Our empirical evaluation on two text classification tasks and two types of label noise: random and input-conditional, shows that our approach can improve classification accuracy, and prevent over-fitting to the noise.
\end{abstract}


\vspace{-2pt}

\settopmatter{printacmref=true}

\maketitle

\vspace{-0.8em}
\section{Introduction}

Training modern ML models requires access to large accurately labeled datasets, which are difficult to obtain due to errors in automatic or human annotation techniques~\cite{wang2018devil,Zlateski2018}.
Recent studies~\cite{zhang2016understanding} have shown that neural models can over-fit on noisy labels and thereby not generalize well.
Human annotations for language tasks have been popularly obtained from platforms like Amazon Mechanical Turk~\cite{10.1145/1837885.1837906}, resulting in noisy labels due to ambiguity of the correct label~\cite{10.1145/3309543}, annotation speed, human error, inexperience of annotator, etc. 
While learning with noisy labels has been extensively studied in computer vision~\cite{journals/corr/ReedLASER14,zhang2018mixup,pmlr-v97-thulasidasan19a}, the corresponding progress in NLP has been limited.
With the increasing size of NLP datasets, noisy labels are likely to affect several practical applications~\cite{Agarwal07howmuch}.

\begin{figure}
    \centering
    
    \includegraphics[width=0.7\linewidth]{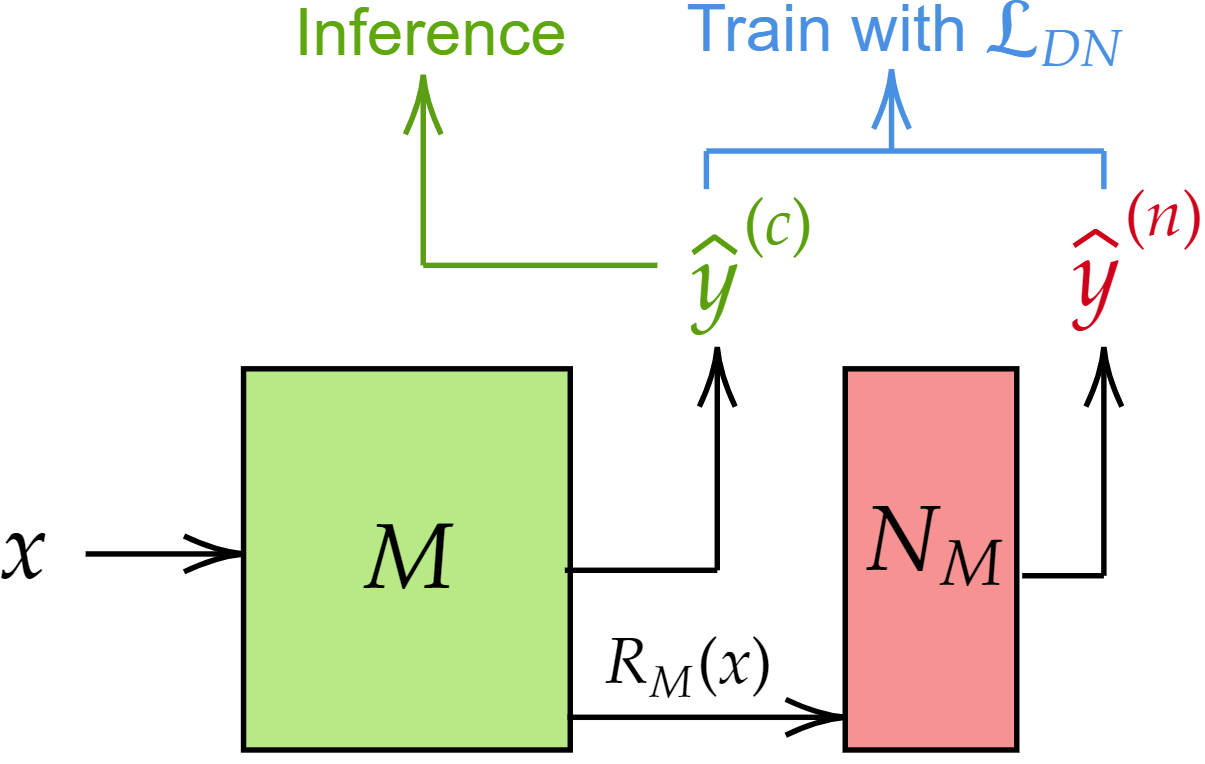} 
    \vspace{-0.5em}
    \caption{
    Illustration of our approach, with an auxiliary noise model $N_M$ on top of the classifier $M$. 
    We jointly train the models using a \textit{de-noising} loss $\mathcal{L}_{DN}$, and use the clean label prediction $\hat y^{(c)}$ during inference. 
    }
    
    \label{fig:model}
    \vspace{-2em}
\end{figure}

In this paper, we consider the problem of text classification, and capture the label noise through an auxiliary noise model (See Fig.~\ref{fig:model}). 
We leverage the finding of learning on clean labels being easier than on noisy labels~\cite{arazo19}, and first fit a two-component beta-mixture model (BMM) on the training losses from the classifier at an early epoch. 
Using this, we assign a probability score to every training sample of having a clean or noisy label. 
We then jointly train the classifier and the noise model by selectively guiding the former's prediction for samples with high probability scores of having clean labels. 
More specifically, we propose a novel de-noising loss having two components: (i) cross-entropy of the noise model prediction with the input label and (ii) cross-entropy of the classifier prediction with the input label, weighted by the probability of the sample having a clean label.
Our formulation constrains the noise model to learn the label noise, and the classifier to learn a good representation for the prediction task from the clean samples.
At inference time, we remove the noise model and use the predictions from the classifier.

\begin{figure*}[t!]
\captionsetup[subfigure]{justification=centering}
    \begin{subfigure}[t]{0.18\linewidth}
        \centering
        \includegraphics[height=0.95in,width=1.15in]{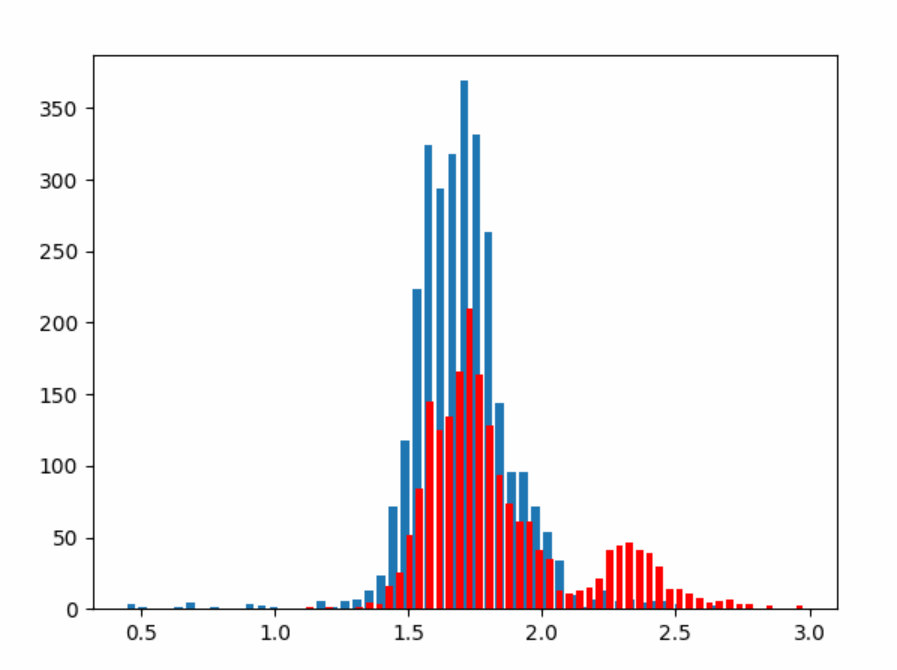}
        \caption{Epoch 1}
    \end{subfigure}%
    ~ 
    \begin{subfigure}[t]{0.18\linewidth}
        \centering
        \includegraphics[height=0.95in,width=1.15in]{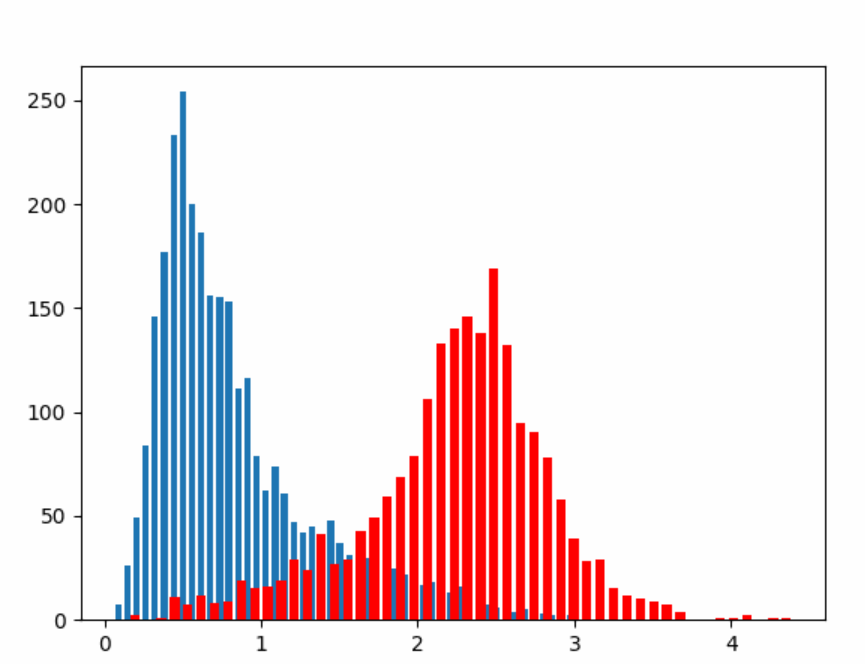}
        \caption{Epoch 9}
    \end{subfigure}
    ~ 
    \begin{subfigure}[t]{0.18\textwidth}
        \centering
        \includegraphics[height=0.95in,width=1.15in]{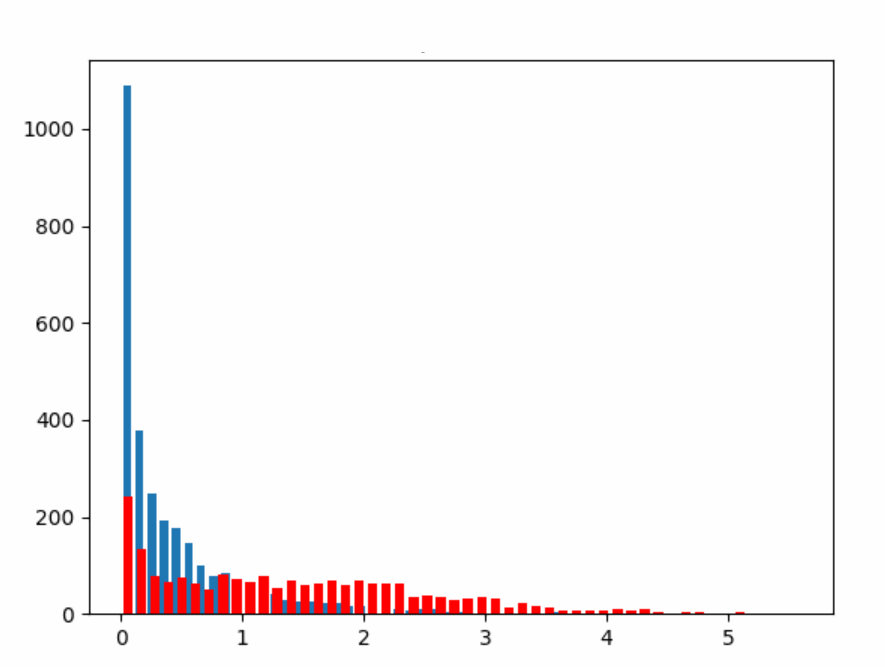}
        \caption{Epoch 30}
    \end{subfigure}
    ~
    \begin{subfigure}[t]{0.38\textwidth}
        \centering
        \includegraphics[height=0.95in, width=2.6in]{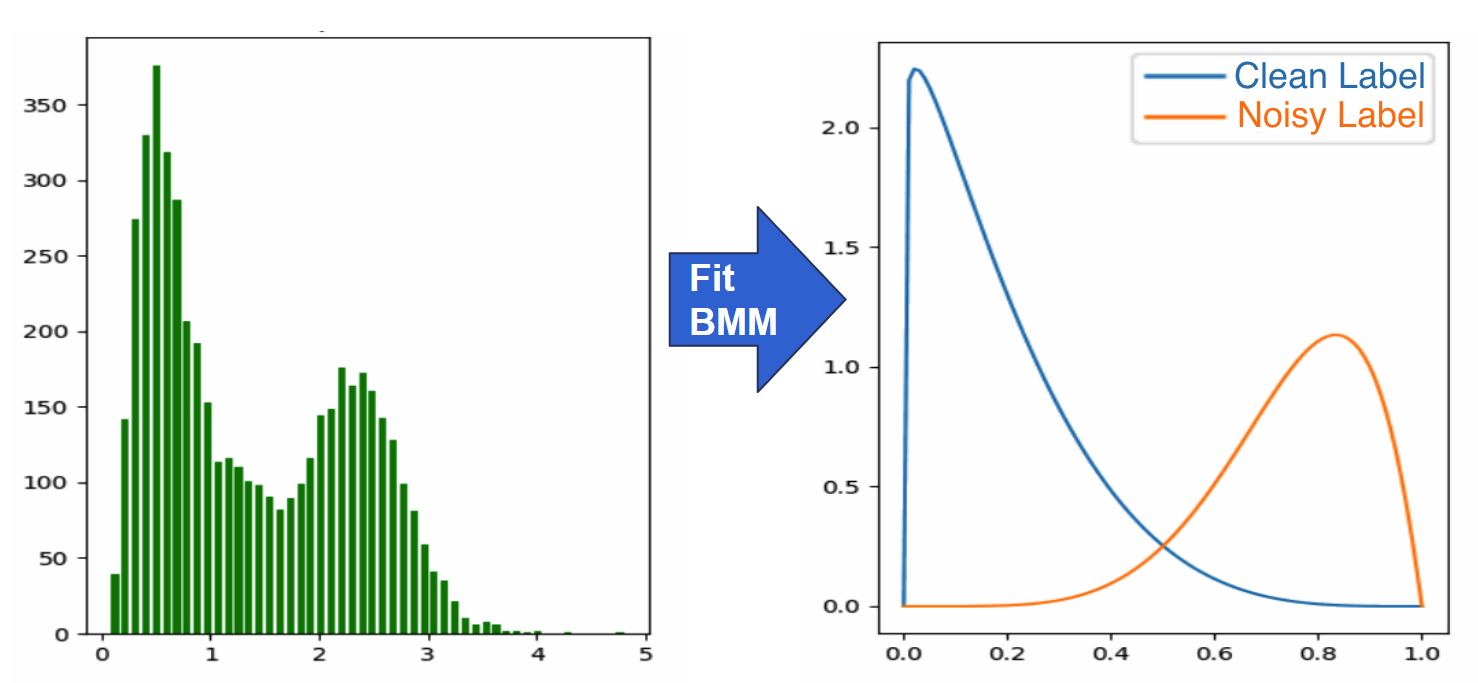}
        \centering{\caption{Fitting a BMM at Epoch 9}}    \end{subfigure}
    \vspace{-0.7em}
    \caption{
    (a), (b) and (c) show the histogram of the training loss from the classifier for the train split (with \textcolor{blue}{Clean}/ \textcolor{red}{Noisy} label) at different epochs (word-LSTM on TREC with 40\% random noise). 
    Initially (see (a)), the losses are high for all data points (both clean and noisy labels).  
    A fully trained model achieves low losses on both clean and noisy data points, indicating over-fitting to the noise, as seen in (c). 
    However, at an early epoch of training, we observe that samples with clean labels have lower losses while those with noisy labels have high losses, leading to the formation of two clusters as seen in (b). 
    We fit a beta-mixture model at this intermediate epoch to estimate the probability of a sample having a clean or noisy label, as shown in (d). 
    }
    \label{fig:hist_epochs}
\end{figure*}

Most existing works on learning with noisy labels assume that the label noise is independent of the input and only conditional on the true label.
Text annotation complexity has been shown to depend on the lexical, syntactic and semantic input features~\cite{joshi-etal-2014-measuring} and not solely on the true label.
The noise model in our formulation can capture an arbitrary noise function, which may depend on both the input and the original label, taking as input a contextualized input representation from the classifier. 
While de-noising the classifier for sophisticated noise functions is a challenging problem, we take the first step towards capturing a real world setting. 

We evaluate our approach on two popular datasets, for two different types of label noise: random and input-conditional; at different noise levels.
Across two model architectures, our approach
results in improved model accuracies over the baseline, while preventing over-fitting to the label noise. 


\section{Related Work}
\label{sec:related_work}
There have been several research works that have studied the problem of combating label noise in computer vision~\cite{Frnay2014ClassificationIT, pmlr-v80-jiang18c,Jiang2019HyperspectralIC} through techniques like bootstrapping~\cite{journals/corr/ReedLASER14}, mixup~\cite{zhang2018mixup}, etc. 
Applying techniques like mixup (convex combinations of pairs of samples) for textual inputs is challenging due to the discrete nature of the input space and retaining overall semantics. 
In natural language processing, 
\citet{Agarwal07howmuch} study the effect of different kinds of noise on text classification, 
\citet{Ardehaly2018} study social media text classification using label proportion (LLP) models, and \citet{Malik_automatictraining} automatically validate noisy labels using high-quality class labels.
\citet{jindal2019effective} capture random label noise via a $\ell_2$-regularized matrix learned on the classifier logits.
Our work differs from this as we i) use a neural network noise model over contextualized embeddings from the classifier, with (ii) a new de-noising loss to explicitly guide learning. It is difficult to draw a distinction between noisy labels, and outliers which are hard to learn from.
While several works perform outlier detection~\cite{goodman-etal-2016-noise,larson-etal-2019-outlier} to discard these samples while learning the classifier, we utilise the noisy data in addition to the clean data for improving performance.

\vspace{-0.8em}

\section{Methodology}
\label{sec:methodoloy}
\myparagraph{Problem Setting}
Let $(X,Y^{(c)}){=} \{ (x_1,y_1^{(c)}),\dots , (x_N,y_N^{(c)})\}$ 
denote clean training samples from a distribution $\mathcal{D}{=}{\mathcal{X}}{\times}{\mathcal{Y}}$.
We assume a function $\mathcal{F}:{\mathcal{X}}{\times}{\mathcal{Y} \rightarrow \mathcal{Y}}$ that introduces noise in labels $Y^{(c)}$.
We apply $\mathcal{F}$ on $(X,Y^{(c)})$ to obtain the noisy training data $(X,Y^{(n)}) = \{(x_1,y_1^{(n)}),\dots, (x_N,y_N^{(n)})\}$.
$(X,Y^{(n)})$ contains a combination of clean samples (whose original label is retained $y^{(n)}{=}y^{(c)}$) and noisy samples (whose original label is corrupted $y^{(n)}{\neq}y^{(c)}$).
Let $(X_{T},Y_{T})$ be a test set sampled from the clean distribution $\mathcal{D}$.
Our goal is to learn a classifier model $\mathcal{M}:\mathcal{X}{\rightarrow}\mathcal{Y}$ trained on the noisy data $(X,Y^{(n)})$, which generalizes well on $(X_{T},Y_{T})$.
Note that we do not have access to the clean labels $Y^{(c)}$ at any point during training.

\vspace{1em}

\myparagraph{Modeling Noise Function $\mathcal{F}$}
We propose to capture $\mathcal{F}$ using an auxiliary \textit{noise model} $N_M$ on top of the classifier model $M$, as shown in Fig. \ref{fig:model}. 
For an input $x$, a representation $R_M(x)$, derived from $M$, is fed to $N_M$. 
$R_M(x)$ can typically be the contextualized input embedding from the penultimate layer of $M$. 
We denote the predictions from $M$ and $N_M$ to be $\hat y^{(c)}$(clean prediction) and $\hat y^{(n)}$(noisy prediction) respectively. 
The clean prediction $\hat y^{(c)}$ is used for inference.

\newcommand{\tuple}[2]{#1 {\small{(#2)}}}
\newcommand{\besttuple}[2]{\textbf{#1} {\small{(#2)}}}

\definecolor{cadmiumgreen}{rgb}{0.0, 0.42, 0.24}
\definecolor{ao(english)}{rgb}{0.0, 0.5, 0.0}

\begin{table*}[t!]
\resizebox{\textwidth}{!}{
\begin{tabular}{cclccccclccccc}
\toprule
\multirow{2}{*}{\textbf{Model}}                                               &                        &  & \multicolumn{5}{c}{\textbf{TREC} (word-LSTM: 93.8, word-CNN: 92.6)}                                                                         &  & \multicolumn{5}{c}{\textbf{AG-News} (word-LSTM: 92.5, word-CNN: 91.5)}                                                                      \\ \cmidrule(lr){4-8} \cmidrule(l){10-14} 
                                                                              & Noise \%               &              & 10            & 20            & 30             & 40             & 50             &           & 10            & 20            & 30             & 40             & 50             \\ \midrule
\multirow{3}{*}{\textbf{\begin{tabular}[c]{@{}c@{}}word\\ LSTM\end{tabular}}} & Baseline               &  & \tuple{88.0}{-0.6} & \tuple{89.4}{\textcolor{red}{-9.6}} & \tuple{83.4}{\textcolor{red}{-19.0}} & \tuple{79.6}{\textcolor{red}{-24.8}} & \tuple{77.6}{\textcolor{red}{-27.2}} &  & \besttuple{91.9}{\textcolor{red}{-1.7}} & \besttuple{91.3}{\textcolor{red}{-1.5}} & \tuple{90.5}{\textcolor{red}{-2.5}}  & \tuple{89.3}{\textcolor{red}{-3.7}}    & \tuple{88.6}{\textcolor{red}{-10.5}} \\
                                                                              & $\mathcal{L}_{DN{-}H}$ &  &  \tuple{92.2}{\textcolor{ao(english)}{-0.6}} & \besttuple{90.2}{\textcolor{ao(english)}{-0.2}}   & \besttuple{88.8}{\textcolor{ao(english)}{-0.4}}  & \tuple{83.0}{-3.6}  & \tuple{82.4}{\textcolor{ao(english)}{0.0}}    &  &  \tuple{91.5}{\textcolor{ao(english)}{-0.1}} & \tuple{90.6}{\textcolor{ao(english)}{-0.1}} & \tuple{90.8}{-0.1}  & \besttuple{90.3}{\textcolor{ao(english)}{0.0}}    & \besttuple{89.0}{\textcolor{ao(english)}{-0.1}}  \\
                                                                              & $\mathcal{L}_{DN{-}S}$ &  &  \besttuple{92.4}{\textcolor{red}{-1.0}} & \tuple{90.0}{-0.2} & \tuple{87.4}{-2}    & \besttuple{83.4}{\textcolor{ao(english)}{-1.0}}  & \besttuple{82.6}{-8.4}  &  &  \tuple{91.8}{-0.3} & \tuple{90.8}{-0.2} & \besttuple{91.0}{\textcolor{ao(english)}{-0.1}}  & \tuple{90.3}{-0.1}  & \tuple{88.6}{-0.1}  \\ \midrule
\multirow{3}{*}{\textbf{\begin{tabular}[c]{@{}c@{}}word\\ CNN\end{tabular}}}  & Baseline               &  &  \tuple{88.8}{\textcolor{red}{-1.4}} & \tuple{89.2}{-1.8} & \tuple{84.8}{\textcolor{red}{-8.0}}  & \besttuple{82.2}{\textcolor{red}{-15.0}} & \tuple{77.6}{\textcolor{red}{-16.0}} &  & \tuple{90.9}{\textcolor{red}{-2.7}} & \tuple{90.6}{\textcolor{red}{-6.2}} & \tuple{89.3}{\textcolor{red}{-10.2}} & \besttuple{89.2}{\textcolor{red}{-17.9}} & \besttuple{87.4}{\textcolor{red}{-25.2}} \\
                                                                              & $\mathcal{L}_{DN{-}H}$ &  &  \tuple{91}{\textcolor{ao(english)}{-0.2}}   & \tuple{90.8}{\textcolor{ao(english)}{-0.2}} & \besttuple{89.4}{\textcolor{ao(english)}{-1.0}}  & \tuple{81.4}{\textcolor{ao(english)}{0.0}}    & \besttuple{81.4}{\textcolor{ao(english)}{-4.8}}  &  &  \besttuple{91.3}{-0.2} & \besttuple{91.0}{-0.4} & \besttuple{90.3}{\textcolor{ao(english)}{-0.3}}  & \tuple{88.3}{\textcolor{ao(english)}{-3.2}}  & \tuple{86.6}{\textcolor{ao(english)}{-3.5}}  \\
                                                                              & $\mathcal{L}_{DN{-}S}$ &  &  \besttuple{92.2}{-1.4} & \besttuple{91.8}{\textcolor{red}{-2.0}} & \tuple{88.8}{-2.8}  & \tuple{77.0}{-2.4}  & \tuple{77.2}{-7.0}  &  &  \tuple{90.9}{\textcolor{ao(english)}{0.0}}   & \tuple{90.4}{\textcolor{ao(english)}{-0.1}} & \tuple{88.7}{-1.1}  & \tuple{86.6}{-3.5}  & \tuple{84.5}{-10.2} \\ \bottomrule
\end{tabular}
}
\caption{
Results from experiments using random noise. Here for A(B): A refers to the 
\textit{Best} model accuracy while B refers to (\textit{Last-Best}) accuracy. 
The models with highest \textit{Best} accuracies are in \textbf{bold}. For each noise $\%$, the least and most reductions in \textit{Last} accuracy are highlighted in \textcolor{ao(english)}{green} and \textcolor{red}{red}. Baseline 
($0\%$ noise) reported beside dataset. 
}
\label{tab:random_results}
\end{table*}

\subsection{Estimating clean/noisy label using BMM}
It has been empirically observed that classifiers that capture input semantics do not fit the noise before significantly learning from the clean samples~\cite{arazo19}.  
For a classifier trained using a cross entropy loss($\mathcal{L}_{CE}$) on the noisy dataset, this can be exploited to cluster the input samples as being clean/noisy in an unsupervised manner. 
Initially the training loss on both clean and noisy samples is large, and after a few training epochs, the loss of majority of the clean samples reduces. 
Since the loss of the noisy samples is still large, this segregates the samples into two clusters with different loss values. 
On further training, the model over-fits on the noisy samples and the training loss on both samples reduces. We illustrate this in Fig.~\ref{fig:hist_epochs}(a)$-$(c).
We fit a 2-component Beta mixture model (BMM) over the normalized training losses ($\mathcal{L}_{CE}(\hat{y}^{(c)},\cdot) \in [0,1]$) obtained after training the model for some warmup epochs $T_0$.
Using a Beta mixture model works better than using a Gaussian mixture model as it allows for asymmetric distributions and can capture the short left-tails of the clean sample losses.
For a sample $(x,y)$ with normalized loss $\mathcal{L}_{CE}(\hat{y}^{(c)},y)=\ell$, the BMM is given by: 
\begin{align}
     &p(\ell) = \lambda_c \cdot p(\ell | \text{clean}) + \lambda_n \cdot p(\ell | \text{noisy}) \nonumber \\
     &p(\ell | \text{clean}) = \frac{\Gamma(\alpha_c + \beta_c)}{\Gamma(\alpha_c)\Gamma(\beta_c)}\ell^{\alpha_c-1}{(1-\ell)}^{\beta_c-1} \nonumber \\
     &p(\ell | \text{noisy}) = \frac{\Gamma(\alpha_n + \beta_n)}{\Gamma(\alpha_n)\Gamma(\beta_n)}\ell^{\alpha_n-1}{(1-\ell)}^{\beta_n-1} \nonumber 
\end{align}
where $\Gamma$ denotes the gamma distribution and $\alpha_{c/n}, \beta_{c/n}$ are the parameters corresponding to the individual clean/noisy Beta distributions. 
The mixture coefficients $\lambda_c$ and $\lambda_n$, and parameters ($\alpha_{c/n}, \beta_{c/n}$) are learnt using the EM algorithm. 
On fitting the BMM $\mathcal{B}$, for a given input $x$ with a normalized loss $\mathcal{L}_{CE}(\hat{y}^{(c)},y)=\ell$, we denote the posterior probability of $x$ having a clean label by:
\[
\mathcal{B}(x) = \frac{\lambda_c \cdot p(\ell | \text{clean})}{\lambda_c \cdot p(\ell | \text{clean}) + \lambda_n \cdot p(\ell | \text{noisy})}
\]
The BMM $\mathcal{B}$ learnt from Fig. \ref{fig:hist_epochs}b is shown in Fig. \ref{fig:hist_epochs}d. 

\setlength{\dbltextfloatsep}{3pt}
\begin{algorithm}[t]
\caption{Training using $\mathcal{L}_{DN-H}$}
\label{alg1}
\begin{algorithmic}
\small
\STATE \textbf{Input:} Train data $(x_i,y_i^{(n)})_{i=1}^{N}$, warmup epochs $T_0$, total epochs $T$, parameter $\beta$, 
classifier $M$, noise model $N_M$
 \FOR{epoch in $\{1,\dots,T_0\}$ }
  \STATE $\hat{y_i}^{(c)} \leftarrow M(x_i) \ \forall \ i \in [N]$\\
    \STATE Train $M$ with $\sum_{i}\mathcal{L}_{CE}(\hat{y_i}^{(c)},y_i^{(n)})$
 \ENDFOR
 \STATE Fit a 2-mixture BMM $\mathcal{B}$ on  $\{\mathcal{L}_{CE}(\hat{y_i}^{(c)},y_i^{(n)})\}_{i=1}^{N} $ \\
 \FOR{epoch in $\{T_0+1,\dots,T\}$ }
    \STATE $\hat{y_i}^{(c)} \leftarrow M(x_i)$, \\      $\hat{y_i}^{(n)} \leftarrow N_M(R_M(x_i)) \ \  \forall \ i \in [N]$\\
    \STATE Train $M,N_M$ with $\mathcal{L}_{DN{-}H} = $ \\ $ \qquad \sum_{i} \big{(} \mathcal{L}_{CE}(\hat{y_i}^{(n)}, y_i^{(n)}) {+}$
    $\beta \cdot \mathbbm{1}[\mathcal{B}(x){>}0.5] \cdot \mathcal{L}_{CE}(\hat{y_i}^{(c)},y_i^{(n)}) \big{)} $
  \ENDFOR
  \STATE \textbf{Return:} Trained classifier model $M$
  \end{algorithmic}
\end{algorithm}

\subsection{Learning $M$ and $N_M$}
We aim to train $M,N_M$ such that when given an input, $M$ predicts the clean label and $N_M$ predicts the noisy label for this input (if $\mathcal{F}$ retains the original clean label for this input, then both $M, N_M$ predict the clean label).
Thus for an input $(x,y)$ having a clean label, we want $\hat y^{(c)}{=}\hat y^{(n)}{=}y$; and for an input $(x,y)$ having a noisy label, we want $\hat y^{(n)}{=}y$ and $\hat y^{(c)}$ to be the clean label for $x$. We jointly train $M,N_M$ using the de-noising loss proposed below:

\begin{align}
    \mathcal{L}_{DN} {=} \ \mathcal{L}_{CE}(\hat{y}^{(n)},y) \ {+}  
    \beta {\cdot} \mathcal{B}(x){\cdot} \mathcal{L}_{CE}({\hat{y}^{(c)}},y) 
\end{align} 

\noindent The first term trains the $M {-} N_M$ cascade jointly using cross entropy between $\hat{y}^{(n)}$ and $y$. 
The second term trains $M$ to predict $\hat{y}^{(c)}$ correctly for samples believed to be clean, weighted by $\mathcal{B}(x)$.
Here $\beta$ is a parameter that controls the trade-off between the two terms.

By jointly training $M,N_M$ with $\mathcal{L}_{DN}$, we implicitly constrain the label noise in $N_M$.
We use an alternative loss formulation by replacing the Bernoulli $\mathcal{B}(x)$ with the indicator $\mathbbm{1}[\mathcal{B}(x){>}0.5]$. For ease of notation, we refer the former (using $\mathcal{B}(x)$) as the soft de-noising loss $\mathcal{L}_{DN{-}S}$ and the latter as the hard de-noising loss $\mathcal{L}_{DN{-}H}$.


Thus we use the following 3-step approach to learn $M$ and $N_M$:
\begin{enumerate}
    \item \textbf{Warmup:} Train $M$ using $\mathcal{L}_{CE}(\hat{y}^{(c)},y)$.
    \item \textbf{Fitting BMM:} Fit a 2-component BMM $\mathcal{B}$ on the $\mathcal{L}_{CE}(\hat{y}^{(c)},y)$ for all $(x,y) \in (X,Y^{(n)})$.
    \item \textbf{Training with $\mathcal{L}_{DN}$:} 
    Jointly train $M$ and $N_M$ end-to-end using $\mathcal{L}_{DN{-}S/H}$. 
\end{enumerate}
We summarize our methodology in Algorithm~\ref{alg1}, when using $\mathcal{L}_{DN-H}$.

\begin{table}[t]
\centering
\small
\resizebox{0.9\linewidth}{!}{
\begin{tabular}{ccccc}
\toprule
\textbf{Dataset} & \textbf{Num. Classes} & \textbf{Train} & \textbf{Validation} & \textbf{Test} \\ \midrule
TREC (\citet{Li:2002:LQC:1072228.1072378})           & 6          & 4949           & 503           &     500         \\ 
AG-News (\citet{10.1145/1062745.1062778})             & 4          & 112000           & 8000           &         7600           \\ \bottomrule
\end{tabular}}
\caption{Summary statistics of the datasets}
\label{tab:datasets}
\vspace{-2.8em}
\end{table}



\begin{table*}[t!]
\resizebox{\textwidth}{!}{
\begin{tabular}{cclccccclccccc}
\toprule
\multicolumn{1}{c}{\multirow{2}{*}{\textbf{Model}}}                           &                        & \multicolumn{1}{c}{} & \multicolumn{5}{c}{\textbf{Token   (How/What) based}}                                                                      & \multicolumn{1}{c}{} & \multicolumn{5}{c}{\textbf{Length based}}                                                                                  \\ \cmidrule(lr){4-8} \cmidrule(l){10-14} 
\multicolumn{1}{c}{}                                                          & Noise \%               & \multicolumn{1}{c}{} & \multicolumn{1}{c}{10} & \multicolumn{1}{c}{20} & \multicolumn{1}{c}{30} & \multicolumn{1}{c}{40} & \multicolumn{1}{c}{50} & \multicolumn{1}{c}{} & \multicolumn{1}{c}{10} & \multicolumn{1}{c}{20} & \multicolumn{1}{c}{30} & \multicolumn{1}{c}{40} & \multicolumn{1}{c}{50} \\ \midrule
\multirow{3}{*}{\textbf{\begin{tabular}[c]{@{}l@{}}word\\ LSTM\end{tabular}}} & Baseline               &                      & \tuple{89.2}{-0.4}         & \tuple{84.4}{-8.2}         & \tuple{77.8}{-10.6}        & \tuple{76}{-17}            & \tuple{71.8}{-15.8}        &                      & \tuple{91.4}{-1.0}         & \tuple{87}{0.6}              & \tuple{82.2}{1.8}          & \tuple{82.4}{-2.6}         & \tuple{74.2}{-3.0}           \\
                                                                              & $\mathcal{L}_{DN{-}H}$ &                      & \besttuple{91.8}{0}            & \tuple{87.4}{-2.2}         & \besttuple{84.2}{0.4}          & \tuple{79}{1}                & \tuple{67.8}{1.4}          &                      & \tuple{91.6}{-0.6}         & \tuple{90.2}{-0.8}         & \besttuple{87.4}{-0.2}         & \besttuple{87.4}{-0.8}         & \besttuple{79}{0}              \\
                                                                              & $\mathcal{L}_{DN{-}S}$ &                      & \tuple{91.8}{0.2}          & \besttuple{90.6}{-1.2}         & \tuple{83.8}{-6.8}         & \besttuple{79.2}{-19.2}          & \besttuple{75.6}{-15.8}          &                      & \besttuple{92}{0.4}            & \besttuple{90.6}{1}            & \tuple{85.4}{0.2}          & \tuple{84}{-2.8}           & \tuple{75}{-3}             \\ \midrule
\multirow{3}{*}{\textbf{\begin{tabular}[c]{@{}l@{}}word\\ CNN\end{tabular}}}  & Baseline               &                      & \tuple{90.4}{-3.6}         & \tuple{83.8}{-1.8}         & \tuple{82.4}{-7.4}         & \tuple{78.8}{-17.2}        & \tuple{52}{1.4}            &                      & \tuple{91}{0}              & \tuple{88}{1.2}            & \tuple{85.2}{-1.2}         & \tuple{82}{-2.6}           & \tuple{73.6}{-1.4}         \\
                                                                              & $\mathcal{L}_{DN{-}H}$ &                      & \tuple{90}{0.8}            & \tuple{86.6}{-3}           & \besttuple{84.4}{-0.6}         & \tuple{80.6}{-4.2}         & \besttuple{74}{-7.6}           &                      & \tuple{90.6}{-0.8}         & \tuple{89.6}{-1}           & \besttuple{87.2}{-0.2}         & \tuple{82.6}{-0.4}         & \besttuple{77}{-6.2}           \\
                                                                              & $\mathcal{L}_{DN{-}S}$ &                      & \besttuple{91.2}{-0.4}         & \besttuple{86.8}{-1.8}         & \tuple{84.2}{-4.2}         & \besttuple{81.8}{-12}          & \tuple{65.2}{-12.4}        &                      & \besttuple{92.8}{-3.6}         & \besttuple{91}{-2.2}           & \tuple{86.8}{-1.4}         & \besttuple{86}{-4}             & \tuple{75.4}{1.8}          \\ \bottomrule
\end{tabular}
}
\caption{
Results from experiments using input-conditional noise on the TREC dataset.
}
\label{tab:inp_trec_results}
\end{table*}

\section{Evaluation} 
\label{sec:eval}

\myparagraph{Datasets}
We experiment with two popular text classification datasets: (i) TREC question-type dataset ~\cite{Li:2002:LQC:1072228.1072378}, and (ii) AG-News dataset ~\cite{10.1145/1062745.1062778} (Table~\ref{tab:datasets}).
We inject noise in the training and validation sets, while retaining the original clean test set for evaluation. 
Note that collecting real datasets with known patterns of label noise is a challenging task, and out of the scope of this work. 
We artificially inject noise in clean datasets, which enables easy and extensive experimentation.

\myparagraph{Models}
We conduct experiments on two popular model architectures: word-LSTM~\cite{Hochreiter:1997:LSM:1246443.1246450} and word-CNN~\cite{kim2014convolutional}.
For word-LSTM, we use a 2-layer BiLSTM with hidden dimension of 150.
For word-CNN, we use 300 kernel filters each of size 3, 4 and 5. 
We use the pre-trained GloVe  embeddings~\cite{pennington2014glove} for initializing the word embeddings for both models.
We train models on TREC and AG-News for 100 and 30 epochs respectively.
We use an Adam optimizer with a learning rate of $10^{-5}$ and a dropout of $0.3$ during training.
For the noise model $N_M$, we use a simple 2-layer feedforward neural network, with the number of hidden units $n_{hidden}=4 {\cdot} n_{input}$.
We choose the inputs to the noise model $R_M(x)$ as per the class of label noise, which we describe in Section~\ref{subsec:random} and \ref{subsec:input_conditional}. 
We conduct hyper-parameter tuning for the number of warmup epochs $T_0$ and $\beta$ using grid search over the ranges of \{6,10,20\} and \{2,4,6,8,10\} respectively.
\vspace{1em}

\myparagraph{Metrics and Baseline}
We evaluate the robustness of the model to label noise on two fronts: (i) how well it performs on clean data, and (ii) how much it over-fits the noisy data. 
For the former, we report the test set accuracy (denoted by \textit{Best}) corresponding to the model with best validation accuracy .
For the latter, we examine the gap in test accuracies between the \textit{Best}, and the \textit{Last} model (after last training epoch). 
We evaluate our approach against only training $M$ (as the baseline), for two types of noise: random and input-conditional, at different noise levels. 
 
\subsection{Results: Random Noise}
\label{subsec:random}
For a specific Noise \%, we randomly change the original labels of this percentage of samples.
Since the noise function is independent of the input, we use logits from $M$ as the input $R(x)$ to $N_M$.
We report the \textit{Best} and (\textit{Last} - \textit{Best}) test accuracies in Table \ref{tab:random_results}. 
From the experiments, we observe that:

(i) $L_{DN-S}$ and $L_{DN-H}$ almost always outperforms the baseline across different noise levels. The performance of $L_{DN-S}$ and $L_{DN-H}$ are similar. We observe that training with $L_{DN{-}S}$ tends to be better at low noise \%, whereas $L_{DN{-}H}$ tends to be better at higher noise \%.
Our method is more effective for TREC than AG-News, since even the baseline can learn robustly on AG-News.

(ii) Our approach using $L_{DN{-}S}$ and $L_{DN{-}H}$ drastically reduces over-fitting on noisy samples (visible from small gaps between \textit{Best} and \textit{Last} accuracies).
For the baseline, this gap is significantly larger, especially at high noise levels, indicating over-fitting to the label noise. 
For example, consider word-LSTM on TREC at 30\% noise: while the baseline suffers a sharp drop of 24.8 points from 79.6\%, the accuracy of the $L_{DN{-}S}$ model drops just 1.0\% from 83.4\%.

\begin{figure}[!t]
        \centering
        \includegraphics[width=0.85\linewidth, height=1.5in]{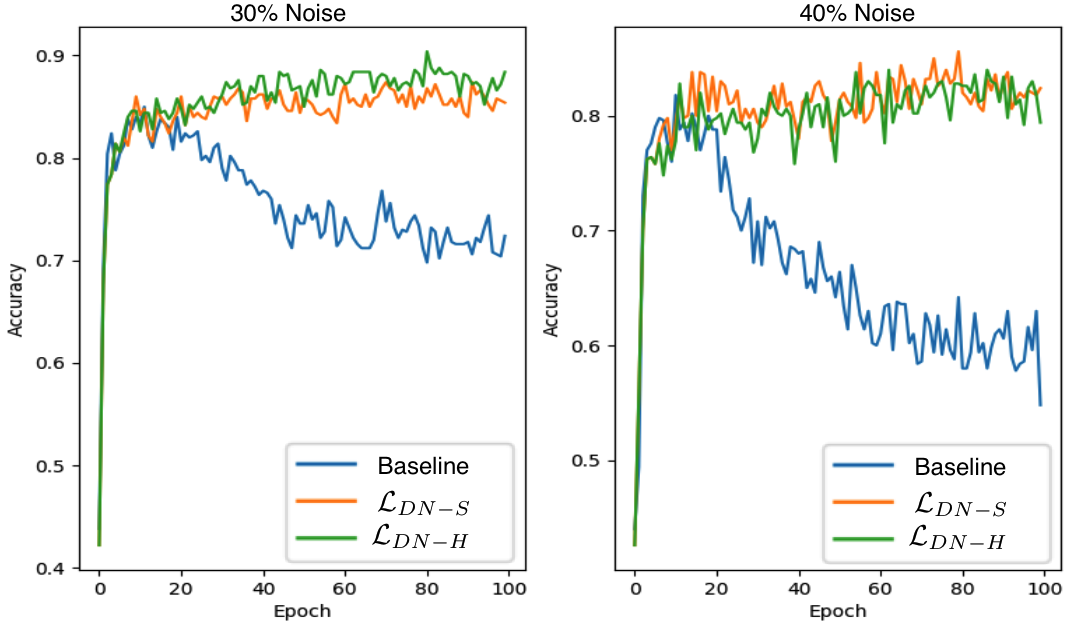}
        \captionof{figure}{Test accuracy across training epochs of word-LSTM model on the TREC dataset with two levels of random noise: $30\%$ and $40\%$. The baseline heavily over-fits on the noise degrading performance, while $L_{DN{-}S}$ and $L_{DN{-}H}$ avoid this.}
        \label{fig:stability}
        \vspace{-1em}
\end{figure}

We further demonstrate that our approach avoids over-fitting, thereby \textit{stabilizing} the model training by plotting the test accuracies across training epochs in Fig.~\ref{fig:stability}. 
We observe that the baseline model over-fits the label noise with more training epochs, thereby degrading test accuracy. 
The degree of over-fitting is greater at higher levels of noise (Fig.~\ref{fig:stability}(b) vs Fig.~\ref{fig:stability}(a)). 
In comparison, our de-noising approach using both $L_{DN-S}$ and $L_{DN-H}$ does not over-fit on the noisy labels as demonstrated by stable test accuracies across epochs. This is particularly beneficial when operating in a few-shot setting where one does not have access to a validation split that is representative of the test split for early stopping.

\begin{table}[t]
\resizebox{\linewidth}{!}{
\begin{tabular}{@{}lcccc@{}}
\toprule
                                                                              & \multicolumn{1}{l}{Noise} & \textbf{AP (7.8\%)}                   & \textbf{Reuters (10.8\%)}              & \textbf{Either (18.6\%)}           \\ \midrule
\multirow{3}{*}{\textbf{\begin{tabular}[c]{@{}l@{}}word\\ LSTM\end{tabular}}} & Baseline                  & \tuple{82.8}{-0.5} & \tuple{85.6}{-0.8} & \tuple{75.7}{-0.4} \\
                                                                              & $\mathcal{L}_{DN{-}H}$    & \tuple{82.7}{0}    & \besttuple{85.7}{-0.1} & \besttuple{76.6}{-0.4} \\
                                                                              & $\mathcal{L}_{DN{-}S}$    & \besttuple{82.8}{0.3}  & \tuple{85.5}{0.1}  & \tuple{76}{-0.1}   \\ \midrule
\multirow{3}{*}{\textbf{\begin{tabular}[c]{@{}l@{}}word\\ CNN\end{tabular}}}  & Baseline                  & \besttuple{83.1}{-0.2} & \tuple{85.7}{0}    & \besttuple{76.6}{-0.9} \\
                                                                              & $\mathcal{L}_{DN{-}H}$    & \tuple{82.4}{0.8}  & \besttuple{86.2}{0}    & \tuple{76.1}{0.1}  \\
                                                                              & $\mathcal{L}_{DN{-}S}$    & \tuple{82.5}{0.5}  & \tuple{86.1}{0.1}  & \tuple{76.4}{0}    \\ \bottomrule
\end{tabular}
}
\caption{
Results from input-conditional noise on AG-News.
}
\label{tab:inp_ag_results}
\vspace{-2em}
\end{table}

\subsection{Results: Input-Conditional Noise}
\label{subsec:input_conditional}
We heuristically condition the noise function $\mathcal{F}$ on lexical and syntactic input features.
We are the first to study such label noise for text inputs, to our knowledge.
For both the TREC and AG-News, we condition $\mathcal{F}$ on syntactic features of the input:
(i) The TREC dataset contains different types of questions. We selectively corrupt the labels of inputs that contain the question words `How' or `What' (chosen based on occurrence frequency). For texts starting with `How' or `What', we insert random label noise (at different levels). We also consider $\mathcal{F}$ conditional on the text length (a lexical feature). More specifically, we inject random label noise for the longest x\% inputs in the dataset.
(ii) The AG-News dataset contains news articles from different news agency sources. We insert random label noise for inputs containing the token `AP', `Reuters' or either one of them.
We concatenate the contextualised input embedding from the penultimate layer of $M$ and the logits corresponding to $\hat{y}^{(c)}$ as the input $R_M(x)$ to $N_M$. 
We present the results in Tables \ref{tab:inp_trec_results} and \ref{tab:inp_ag_results}.

On TREC, our method outperforms the baseline for both the noise patterns we consider.
For the question-length based noise, we observe the same trend of $L_{DN{-}H}$ outperforming $L_{DN{-}S}$ at high noise levels, and vice-versa. 
On AG-News, the noise \% for inputs having the specific tokens 'AP' and 'Reuters' are relatively low, and our method performs at par or marginally improves over the baseline performance. 
Interestingly, the input--conditional noise we consider makes effective learning very challenging, as demonstrated by significantly lower \emph{Best} accuracies for the baseline model than for random noise. 
As the classifier appears to overfit to the noise very early during training, we observe relatively smaller gaps between Best and Last accuracies.
Compared to random noise, our approach is less efficient at alleviating the \emph{(Best-Last)} accuracy gap for input-conditional noise. These experiments however reveal promising preliminary results on learning with input-conditional noise.

\vspace{-0.6em}
\section{Conclusion}
\label{sec:conclusion}

We have presented an approach to improve text classification when learning from noisy labels by jointly training a classifier and a noise model using a de-noising loss.
We have evaluated our approach on two text classification tasks. 
demonstrate its effectiveness through an extensive evaluation.
Future work includes studying more complex $\mathcal{F}$ for other NLP tasks like language inference and QA.

\bibliographystyle{ACM-Reference-Format}
\balance
\bibliography{refs}



\end{document}